\documentclass[10pt,twocolumn,letterpaper]{article}

\usepackage{cvpr}              

\usepackage{graphicx}
\usepackage{amsfonts}
\usepackage{amsmath}
\usepackage{amssymb}
\usepackage{diagbox}
\usepackage{multirow}
\usepackage{graphics}
\usepackage{epstopdf}
\usepackage{tabularx}
\usepackage{epsfig}
\usepackage{times}
\usepackage{mathrsfs}
\usepackage{bm}
\usepackage{algorithm, algorithmic}
\usepackage{mathdots}
\usepackage{microtype} 
\usepackage{array}
\usepackage{setspace}
\usepackage{threeparttable}
\usepackage{url}
\usepackage[colorlinks,linkcolor=red]{hyperref}

\usepackage[capitalize]{cleveref}
\crefname{section}{Sec.}{Secs.}
\Crefname{section}{Section}{Sections}
\Crefname{table}{Table}{Tables}
\crefname{table}{Tab.}{Tabs.}

\def\cvprPaperID{6247} 
\def\confName{CVPR}
\def\confYear{2023}

\begin{document}

\title{Physically Adversarial Infrared Patches  with Learnable Shapes and Locations}

\author{Xingxing Wei\thanks{Corresponding author}, \quad Jie Yu,\quad Yao Huang\\
Institute of Artificial Intelligence, Beihang University, Beijing, 100191, China\\
{\tt\small xxwei@buaa.edu.cn, sy2106137@buaa.edu.cn, y\_huang@buaa.edu.cn}
}
\maketitle

\begin{abstract}
   Owing to the extensive application of infrared object detectors in the safety-critical tasks, it is necessary to evaluate their robustness against adversarial examples in the real world. However, current few physical infrared attacks are complicated to implement in practical application because of their complex transformation from digital world to physical world.  To address this issue, in this paper, we propose a physically feasible infrared attack method called "adversarial infrared patches". Considering the imaging mechanism of infrared cameras by capturing objects' thermal radiation, adversarial infrared patches conduct attacks by attaching a patch of thermal insulation materials on the target object to manipulate its thermal distribution. To enhance adversarial attacks, we present a novel aggregation regularization to guide the simultaneous learning for the patch' shape and location on the target object. Thus, a simple gradient-based optimization can be adapted to solve for them. We verify adversarial infrared patches in different object detection tasks with various object detectors. Experimental results show that  our method achieves more than 90\% Attack Success Rate (ASR) versus the pedestrian detector and vehicle detector in the physical environment, where the objects are captured in different angles, distances, postures, and scenes. More importantly, adversarial infrared patch is easy to implement, and it only needs 0.5 hours to be constructed in the physical world, which verifies its effectiveness and efficiency. 
\end{abstract}

\section{Introduction}
\label{sec:intro}

Deep Neural Networks (DNNs) have shown promising performance in various vision tasks, including object detection \cite{redmon2018yolov3}, classification \cite{krizhevsky2012imagenet}, face recognition \cite{schroff2015facenet}, and autonomous driving \cite{sandler2018mobilenetv2}.  However, it is typically known that DNNs are vulnerable to adversarial examples \cite{goodfellow2014explaining, xiao2018generating, carlini2017towards}, i.e., the human-imperceptible perturbed inputs can fool the DNNs-based system to give wrong predictions. Moreover, these adversarial examples can be exploited in the physical world. In such cases, a widely used technique is called adversarial patches \cite{brown2017adversarial, eykholt2018robust, yang2020patchattack,wei2022physically}, which have been successfully applied to traffic sign detection  by generating a carefully designed sticker \cite{eykholt2018robust,wei2021generating}, or face recognition by adding specific textures on eyeglass frames \cite{sharif2016accessorize, wei2022adversarial}. The success of adversarial patches raises the concerns because of their great threat to the deployed DNN-based systems in the real world.

\begin{figure}[t]
\begin{center}
\includegraphics[width=0.4\textwidth]{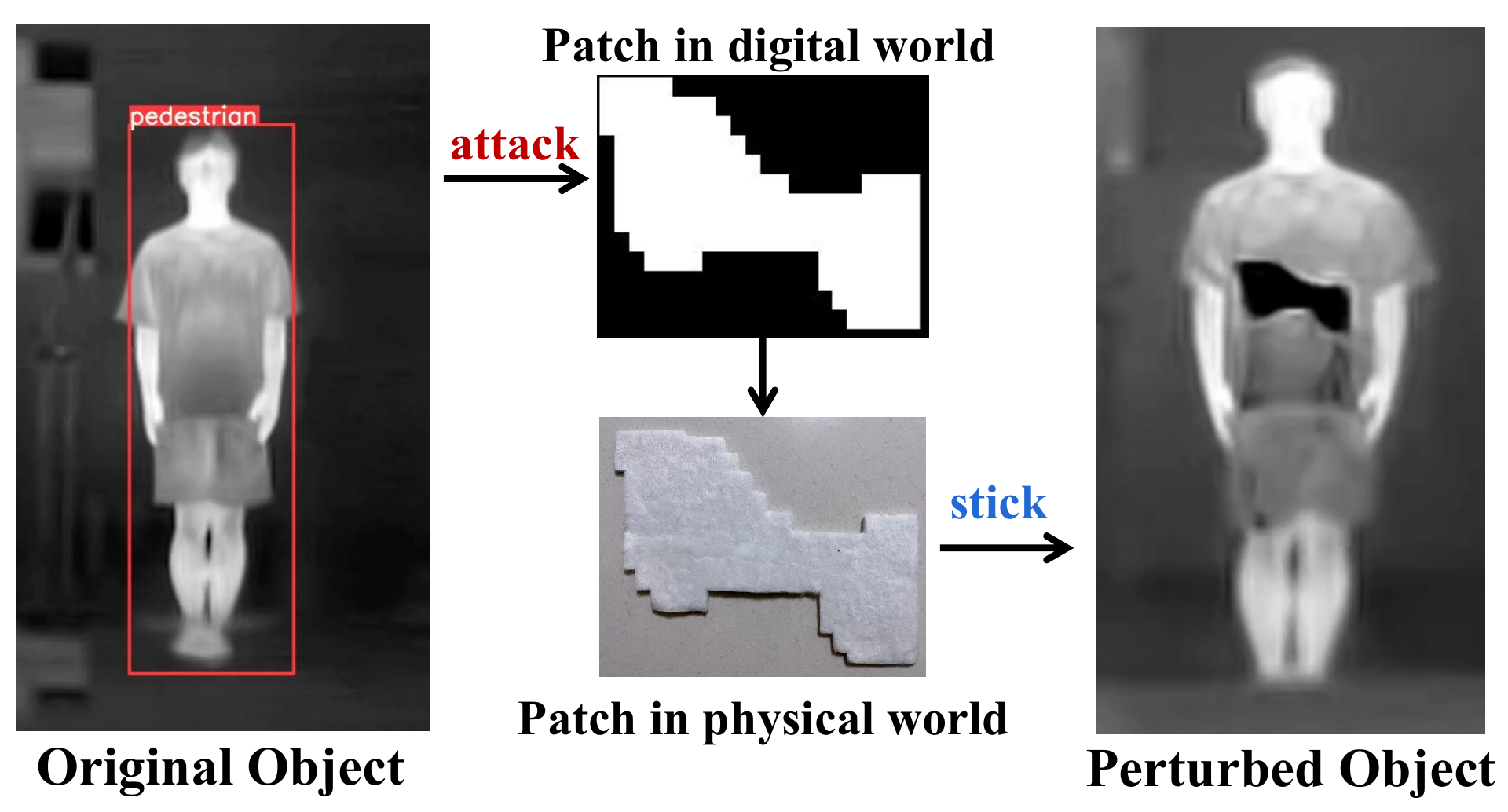}
\end{center}
\vspace{-0.5cm}
\caption{The generation process of adversarial infrared patches. We see the pedestrian cannot be detected after the infrared patches are pasted on the pedestrian in the physical world.}
\label{fig:odot}
\vspace{-0.5cm}
\end{figure}

Nowadays, famous for its strong anti-interference ability in the severe environment, object detection in the thermal infrared images has
been widely used in many safety-critical tasks such as security surveillance \cite{suard2006pedestrian}, remote sensing \cite{weng2009thermal}, etc. Consequently, it is necessary to evaluate the physical adversarial robustness of infrared object detectors. However, the aforementioned adversarial patches cannot work well in the infrared images because they depend on the adversarial perturbations generated from the view of RGB appearance. These perturbations cannot be captured by infrared cameras, which perform the imaging by encoding the objects' thermal radiation\cite{vollmer2021infrared}.  Although few recent works have been proposed to address this issue, they have their own limitations. For example, adversarial bulbs based on a cardboard of alight small bulbs \cite{zhu2021fooling} are complicated to implement in the real world, and are also not stealthy because they produce heat source. Adversarial clothing \cite{zhu2022infrared} based on a large-scale QR code improves the stealthiness by utilizing the thermal insulation material to cover the object's thermal  radiation, but still has complex transformations from digital to physical world, making it not easy to implement in the real world.

In this paper, we propose a physically stealthy and easy-to-implement infrared attack method called ``adversarial infrared patches". Considering the imaging mechanism of infrared cameras by capturing objects' thermal radiation, we also attach the thermal insulation materials on the target object to manipulate its thermal distribution to ensure the stealthiness. But different from adversarial clothing \cite{zhu2022infrared} via the complex QR code pattern, we utilize a simple patch to crop the thermal insulation material, and then adjust the patch's shape and location on the target object to conduct attacks. Compared with  adversarial RGB perturbations, the changes of shapes and locations of the thermal patch can be accurately captured by infrared cameras, which helps perform an effective attack. 
\begin{figure}[t]
\begin{center}
\includegraphics[width=0.4\textwidth]{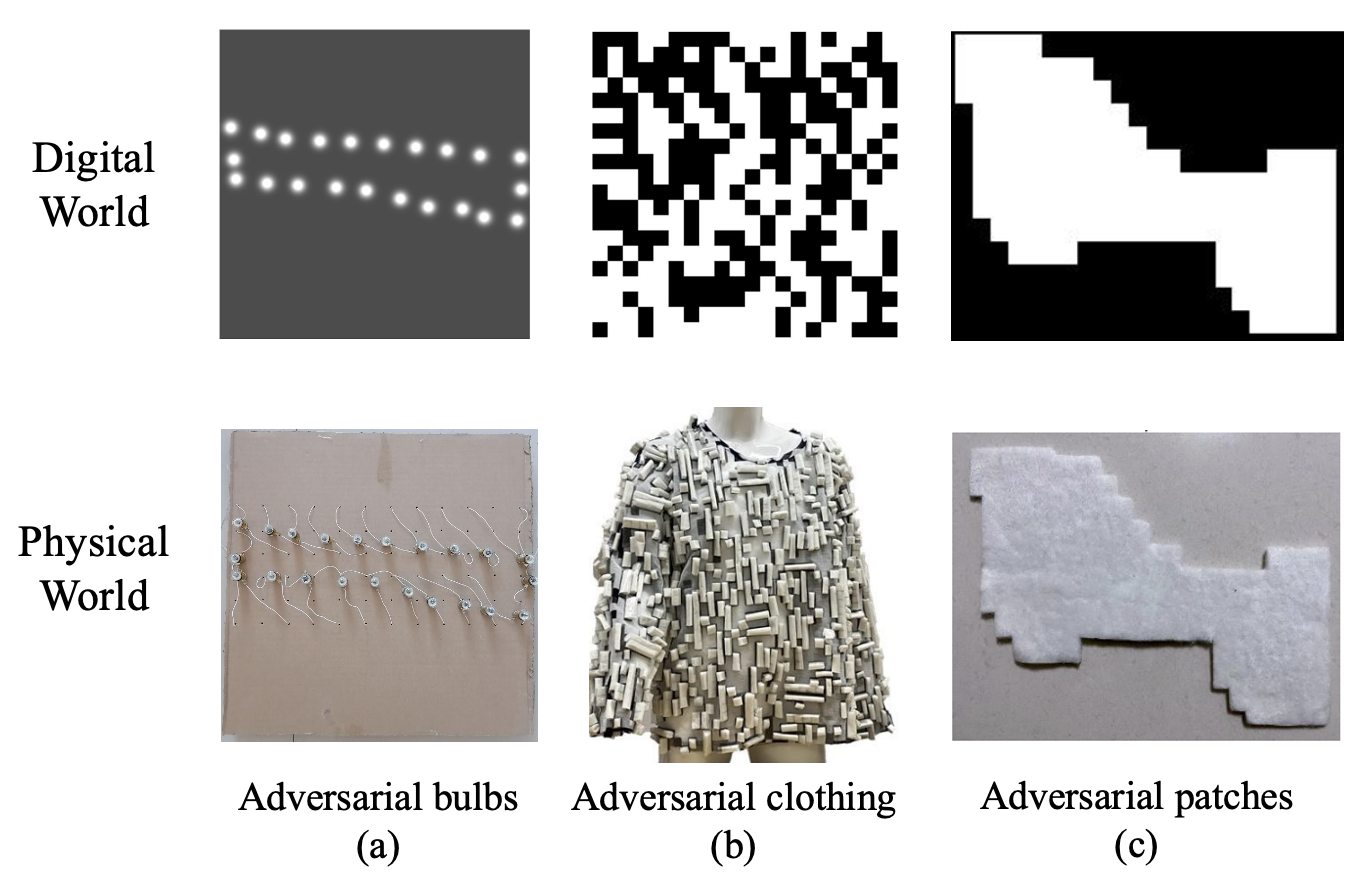}
\end{center}
\vspace{-0.7cm}
\caption{The comparison between different infrared attacks.}
\label{fig:comparion1}
\vspace{-0.5cm}
\end{figure}

However, the shape and location are two kinds of different variables, it is challenging to directly optimize them via unified gradients. For that, we present a novel aggregation regularization to guide the simultaneous learning for the patch' shape and location on the target object. Specifically, an aggregation degree is proposed to quantify how close one pixel's neighbours are to being a clique. By combining this metric with the attack goal, the object's pixels needing to be covered by the thermal insulation material will automatically be gathered to form a valid shape on the available location of the object.  In this way, we can adapt a simple gradient-based optimization to solve for the optimal shape and location of the patch.  An example of our adversarial infrared patch against the pedestrian detector  is shown in Figure \ref{fig:odot}, and a comparison with the existing physical infrared attacks is given in Figure \ref{fig:comparion1}. We see that adversarial infrared patch is simpler than other methods in the digital world, and we just need to crop the thermal insulation materials according to the learned shape, and then paste the patch on the learned location of the pedestrian, which is more easy-to-implement than other methods in the real world.

Our contributions can be summarized as follows:
\begin{itemize}
\setlength{\itemsep}{0pt}
\item We propose the novel ``adversarial infrared patches", a physically stealthy and easy-to-implement attack method for infrared object detection. Instead of generating adversarial perturbations, we perform attacks by learning the available shape and location on the target object. Owing to this careful design, adversarial infrared patches are easier to implement in the physical world than the existing methods. 
\item We design a novel aggregation regularization to
guide the simultaneous learning for the patch’ shape and
location on the target object. Thus, a simple gradient-based
optimization can be adapted to solve for the optimal shape and location of the patch.

\item We verify the adversarial infrared patches in the pedestrian detection task from both the digital and physical world. Experiments show that adversarial infrared patches can work well in various angles, distances, postures, and scenes, and achieve competitive attacking performance with the SOTA infrared attack while only costing five percent of their time to construct physical adversarial examples. We also extend our method to the vehicle detection task to verify its generalization.  
\end{itemize}

\section{Related Works}

Adversarial examples exist widely in various fields.
Recently, adversarial examples in infrared images are explored. Edwards, \emph{et al.}\cite{edwards2020study} investigate the performance of adversarial attack in ship detection under thermal infrared images. \cite{osahor2019deep} explores how to generate visually imperceptible adversarial infrared examples where an object detector based on DNN cannot detect the targets in the image. These methods generate perturbation by changing pixel values within the infrared image, and thus are not applicable in the physical world. 


To address this issue. Zhu, \emph{et al.} \cite{zhu2021fooling} make the first attempt to generate a physical adversarial example using a set of small bulbs. These small bulbs change the infrared radiation distribution of the object via simulating the extra heat sources. Although the decorated bulbs can fool infrared pedestrian detectors, they are not stealthy and easy-to-implement. The reason for poor stealthiness is that they produce heat source rather than cover them, which is not suitable for some scenes that need to hide heat sources from the thermal detector. The reason for poor implementation is that they need to carefully design a circuit board to light the small bulbs, which is complicated in real world. Afterwards, Zhu, \emph{et al.}\cite{zhu2022infrared}  propose adversarial clothing which wraps the whole body to fool the infrared detector at different angles. They aim to design the  QR code pattern in the clothing and then transfer this 2D clothing to 3D clothing. The main issue is the complex operations from digital world to physical world. Specifically, they first print the QR code  on a large-scale cloth. Then, a tailor is hired to make the cloth into a piece of clothing. Finally, the infrared material is cropped into blocks and pasted on the black area of the clothes. 

Our method is different from the existing methods as follows: (1) Compared with \cite{zhu2022infrared}, our adversarial infrared patches just need to design the shape and location of infrared patches rather than the complex patterns. Therefore, it is easy and efficient to implement in the physical world. (2) Because the thermal insulation materials can reduce the pedestrian's thermal radiation rather than enhance it like \cite{zhu2021fooling}, our method is stealthy in the practical scene.  (3) \cite{zhu2022infrared} and \cite{zhu2021fooling} only verify their methods in the pedestrian detection task, while we verify the proposed methods in both the pedestrian detection and vehicle detection, showing the good generalization.

\section{Methodology}
For simplicity, we introduce our method in the pedestrian detection task. The extension to other task is straightforward. 

\subsection{Problem Formulation}
\label{sec:problem}
In the infrared pedestrian detection, suppose $f(\cdot)$ denotes the pedestrian detector with parameter $\theta$, and $f(\bm{x};\theta)$ denotes the output prediction when  a clean thermal image $\bm{x}\in\mathbb{R}^{h\times w}$ is given. For simplicity, we here assume there is only one pedestrian in $\bm{x}$. Thus $y=f(\bm{x};\theta)$, where $y$ is the predictions including the pedestrian's candidate bounding boxes $\{b_i|i=1,...,n\}$ and confidence scores $\{s_i|i=1,...,n\}$. The bounding box with the highest score is regarded as the final detected pedestrian.  

The goal of adversarial examples $\bm{x}_{adv}$ is to prevent the pedestrian detector from detecting the pedestrian. Inspired by \cite{song2018physical}, we minimize the maximum confidence score within all the bounding boxes until it falls below the detection threshold. Thus the pedestrian disappears in the thermal image. This disappear attack loss  is defined as follows:
\begin{equation}
\mathcal{L}_{attack}(f(\bm{x}_{adv})) = \max_{i\in \{1,...,n\}}(s_i).
\label{eq:goal2}
\end{equation}

In this paper, we propose the adversarial infrared patch to construct the adversarial example. It  can be defined as:
\begin{equation}\label{eq:x_adv}
\bm{x}_{adv}=\bm{x}\odot (\bm{1}-\bm{M})+\bm{\hat{x}}\odot \bm{M}, 
\end{equation}
where $\odot$ is Hadamard product, $\bm{M}\in\{0,1\}^{h\times w}$ is a mask matrix used to constrain the shape and location of the infrared patches on the target object, $\bm{\hat{x}}\in \mathbb{R}^{h\times w}$ denotes a cover image used to manipulate  $\bm{x}\in\mathbb{R}^{h\times w}$. The infrared patches can be described as all the region where $\bm{M}_{jk}=1$. It is clear that the location and shape of infrared patches depend on the mask $\bm{M}$, and the content of infrared patches depends on the cover image $\bm{\hat{x}}$. 

In the real world, $\bm{\hat{x}}$ can not be designed freely just as the content of adversarial patches in light visible area \cite{pautov2019adversarial,komkov2021advhat}. It depends on the thermal property of infrared patch material which mainly includes thermal reflectivity and thermal emissivity \cite{vollmer2021infrared}. Since infrared patch is generally made of the same material, the values of $\bm{\hat{x}}$ are usually the same. Therefore, once the material used to make infrared patches has been picked, the values of $\bm{\hat{x}}$ can be determined. And then we just need to optimize the mask $\bm{M}$ to generate the infrared patches to find the adversarial examples.

\subsection{Optimization for Mask}
\label{sec:opt}

We here utilize a gradient-based optimization to solve for the optimal $\bm{M}$. To determine the patch's shape and location, the solved $\bm{M}$ should have the following properties: (1) The pixels with ones in $\bm{M}$ should stick together, and thus they can form a valid shape on an available location. (2) The pixel values in mask $\bm{M}$ should tend to one or zero as possible. However, directly solving Eq.(\ref{eq:goal2}) will lead to a continuous change ranging from 0 to 1 to the value in $\bm{M}$, and the pixels with high values will scatter in different location, which cannot meet the above goal. So we add the aggregation regularization and binary regularization to Eq.(\ref{eq:goal2}) to regularize the pixel values in the optimization, and achieve the above properties, respectively. 

\subsubsection{Aggregation regularization}
As we known, the patch areas are well-shaped, i.e., the pixels with value ones in $\bm{M}$ should be clustered. To measure this aggregation of pixels in  mask $\bm{M}$, we introduce local clustering coefficient \cite{holland1971transitivity,watts1998collective}, which quantifies how close one point's neighbours are to being a clique. For an undirected graph, the local clustering coefficient is defined as:
\begin{equation}
    C_i = \frac{2|\{e_{jk}: v_j,v_k \in L_i\}|}{k_i(k_i-1)}
    \label{eq:connected},
\end{equation}
where $L_i$ represents the set of immediately connected vertices with vertex $v_i$. $v_j$ and $v_k$ are two vertices that belong to $L_i$. $e_{jk}$ denotes the edge  connecting vertex $v_j$ with vertex $v_k$, $k_i$ denotes the number of neighbours of vertex $v_i$. $\vert\cdot\vert$ denotes the number of edges. Please refer to \cite{holland1971transitivity,watts1998collective} for details.

\begin{figure}[t]
\begin{center}
\includegraphics[width=1\linewidth]{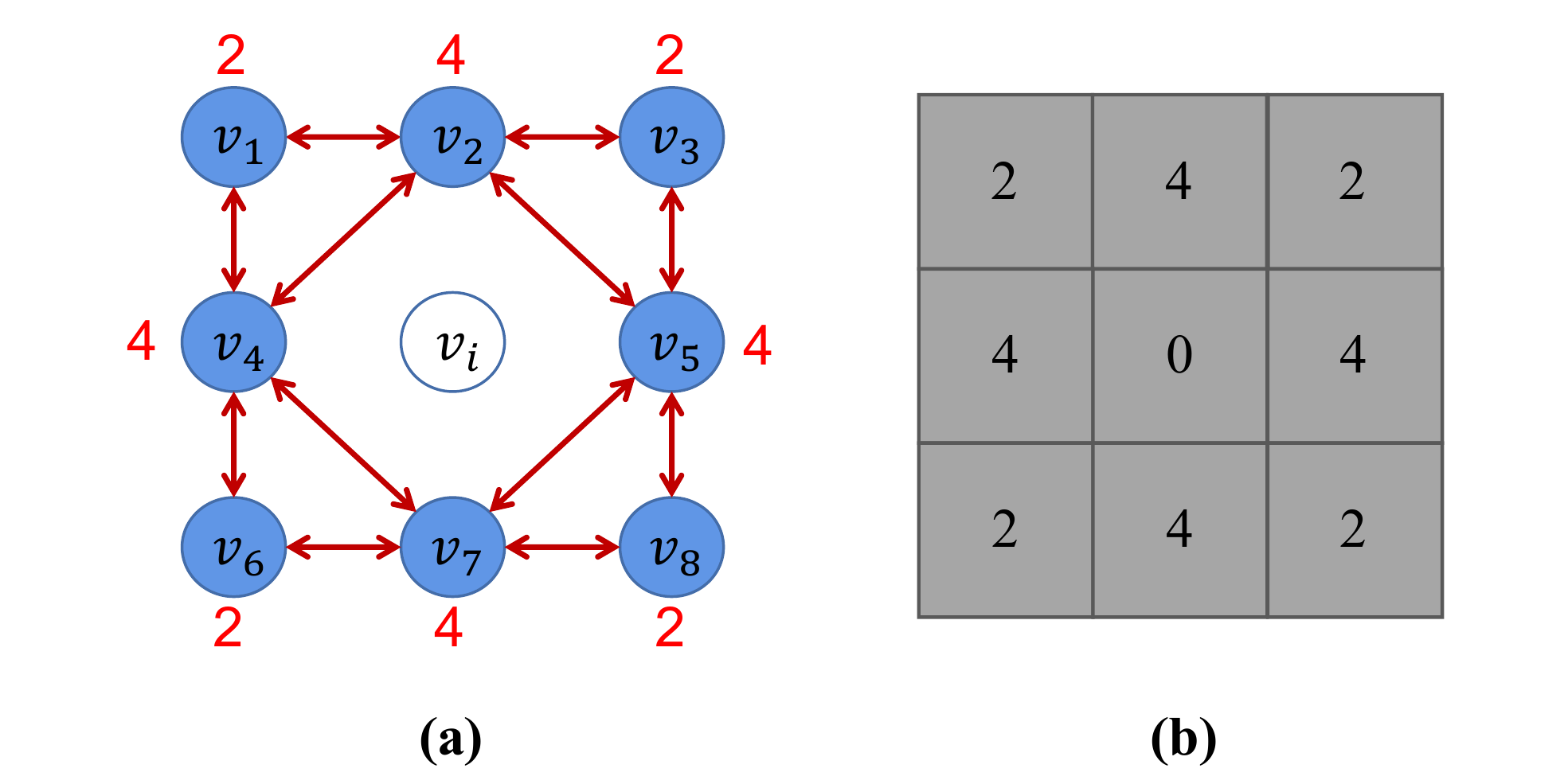}
\end{center}
\vspace{-0.5cm}
\caption{The left picture shows the connection among all the 8 neighbors of the center vertex. The right picture represents the kernel $\mathcal{K}$, where the value of the position represents the number of its immediately connected neighbours in the 8 vertices.}
\label{fig:kernel}
\vspace{-0.3cm}
\end{figure}

Taking Figure \ref{fig:kernel} as an example, we aim to compute the aggregation degree for the center white point $v_i$. Thus, there are 8 immediately connected vertices around it, which is illustrated by the blue points.  According to the definition of $e_{jk}$ in Eq.(\ref{eq:connected}), we plot all the edges with red lines in Figure \ref{fig:kernel}. For simplicity, we compute the edge number according to each vertex. Thus, the edge number for each vertex is given in the red text. If we define a kernel $\mathcal{K}$ using the edge number for each vertex, i.e. the matrix in Figure \ref{fig:kernel} (b), then the edge number $|\{e_{jk}: v_j,v_k \in L_i\}|$ in Eq.(\ref{eq:connected}) can be directly computed by $(\mathcal{K} * \bm{I}_{3\times 3})/2$, where $\bm{I}_{3\times 3}$ denotes a matrix full of one and has the same size as $\mathcal{K}$, and $*$ is the convolution operation. Thus, Eq.(\ref{eq:connected}) is reformulated as:
\begin{equation}
    C_i = \frac{\mathcal{K}* \bm{I}_{3\times 3}}{k_i(k_i-1)}
    \label{eq:connected1}.
\end{equation}
In the example of Figure \ref{fig:kernel} (a), there are 8 neighbour vertices around the center white vertex, therefore, $k_i=8$ here. 

However, in some situations, the vertex may have a value ranging from 0 to 1, rather than 0 or 1. Therefore, we design a soft version for Eq.(\ref{eq:connected1}). In Figure \ref{fig:kernel}, a decay factor is assigned to each vertex according to its own value and the values of its neighbour vertex. The decay factor for the vertex $v_j$ is defined as follows:
\begin{equation}\label{eq:decay}
  \alpha_j= V_j\times \frac{1}{|L_j|}\sum_{v_k\in L_j} V_k,
\end{equation}
where $V_j$ is the vertex $v_j$'s value, $L_j$ is the set of vertex connected with $v_j$, and $V_k$ denotes the vertex's value located in $L_j$. $|L_j|$ is the size of $L_j$. For example, for the vertex $v_2$ in Figure.\ref{fig:kernel}, there are four neighbour vertex, thus $L_2=\{v_1, v_3,v_4, v_5\}$, and $|L_2|=4$. For the vertex $v_1$, there are two neighbour vertex, thus $L_1=\{v_2, v_4\}$, and $|L_1|=2$. After computing $\alpha_j$ to obtain the decay factor matrix $\bm{A}_{3\times 3}$, we replace $\bm{I}_{3\times 3}$ with $\bm{A}_{3\times 3}$ as follows:
\begin{equation}
    C_i = \frac{\mathcal{K}* \bm{A}_{3\times 3}}{k_i(k_i-1)}= \frac{\mathcal{K}* \bm{A}_{3\times 3}}{56}.
    \label{eq:connected2}
\end{equation}

In our task, we use Eq.(\ref{eq:connected2}) to compute the aggregation degree $C_{jk}$ for each element $M_{jk}$ in $\bm{M}$. We have so far defined the $C_{jk}$ of $M_{jk}$,  which has nothing to do with the $M_{jk}$ itself. To take the value of $M_{jk}$ into consideration, we multiply the $C_{jk}$ by $M_{jk}$. Finally, we take the average of all the $C_{jk}$ into consideration, thus the aggregation regularization $\mathcal{L}_{agg}(\bm{M})$ and aggregation degree $C(\bm{M})$ is defined:
\begin{equation}\label{eq:AC}
     \mathcal{L}_{agg}(\bm{M}) = -C(\bm{M}) = -\frac{1}{hw} \sum_{j=1}^h\sum_{k=1}^w C_{jk}\cdot M_{jk},
\end{equation}
where a larger $C(\bm{M})$ means a better aggregation degree.

Obviously, all the operations of the calculation of $\mathcal{L}_{agg}$ are 
differentiable which can be applied in the gradient-based optimization method.
Our goal is to decrease $\mathcal{L}_{agg}$ to make the cluster of infrared patches large enough so that the points in mask $\bm{M}$ gather stick on an available location.

\subsubsection{Binary regularization}
As mentioned before, $\bm{M}$ is a binary matrix. To meet this goal,  we use MSE (Mean Square Error) to construct the regularization as follows: 
\begin{equation}
    \mathcal{L}_{MSE}(\mathcal{H}(\bm{M}), \bm{I}) = \frac{1}{hw}\sum_{j=1}^h\sum_{k=1}^{w} (\mathcal{H}(M_{jk})-1)^2,
    \label{eq:mse}
\end{equation}
where $\bm{I}$ is a matrix whose each element is 1. The function $\mathcal{H}$ here is akin to ReLU function. $\mathcal{H}(M_{jk})=1$ if $M_{jk}\le V_{thre}$ and $\mathcal{H}(M_{jk})=M_{jk}$ if $M_{jk}\geq V_{thre}$. In the ReLU function, the derivative of point whose value is equal to zero is set as zero, while in the function $\mathcal{H}$, the derivative of point whose value equals to $V_{thre}$ is also set as zero, and thus we can make  $\mathcal{H}$ differentiable like ReLU function. 
By minimizing Eq.(\ref{eq:mse}), we only measure the distance between one and the pixels' values in $\bm{M}$ which are greater than threshold $V_{thre}$.

In addition, because the patch should be a localized region on the target object, we add the sparse restriction on $\bm{M}$.  Since $L_0$-norm is non-differentiable, we replace it with $L_1$-norm. By minimizing $\vert\vert\bm{M}\vert\vert_{1}$, we can select as sparse pixels as possible to make the pedestrian disappear in the image. 
The final regularization is defined as follows:
\begin{equation}
    \mathcal{L}_{binary}(\bm{M}) = \vert\vert \bm{M} \vert\vert_1 + \alpha \cdot \mathcal{L}_{MSE}(\mathcal{H}(\bm{M}), \bm{I}),
\end{equation}
where $\alpha$ is a weighted factor. Our goal is to minimize $\mathcal{L}_{binary}$  so that we optimize the sparse and binary $\bm{M}$.

\subsubsection{Total loss function}
Now, we combine three losses mentioned above as our total loss, and keep balance among the losses with weighted factors $\lambda_1$ and $\lambda_2$. The final objective function is given by:
\begin{equation}\label{eq:totalloss}\begin{aligned}
    \mathcal{L}_{obj} = \mathcal{L}_{attack}(f(\bm{x}_{adv})) + \lambda_1\mathcal{L}_{binary}(\bm{M}) + \lambda_2 \mathcal{L}_{agg}(\bm{M})
\end{aligned}
\end{equation}
\begin{equation}
    \bm{M}^*=\arg\min \limits_{\bm{M}} \mathcal{L}_{obj}(f(\bm{x}_{adv})),
\end{equation}
where the weighted factors can be obtained by experience to meet our requirements. In the implementation,  to ensure the infrared patches to attach on the target object, they should be restricted to the region of the target object in the infrared image. In detail, given an available binary $\bm{M}_{obj}$ as the object mask to locate the target object, we can process $\bm{M}=\bm{M}\cdot \bm{M}_{obj}$ in Eq.(\ref{eq:x_adv}) to perform the following optimization.  $\bm{M}_{obj}$ can be obtained via object detection or segmentation. 

\subsubsection{Optimization}
To solve Eq.(\ref{eq:totalloss}), we use the gradient descent algorithm with momentum\cite{qian1999momentum}. Let $t$ denote the $t$-th iteration,  $\bm{x}_t^{adv}$, $M_t$ denote the adversarial image and infrared mask respectively in the $t$-th iteration, then:
\begin{equation}
    \bm{x}_{t+1}^{adv} = \bm{x}\odot (1-\bm{M}_{t+1})+\bm{\hat{x}}\odot \bm{M}_{t+1},
    \label{eq:advget}
\end{equation}
\begin{equation}\label{eq:upgr}
    \bm{M}_{t+1} = (\bm{M}_{t} - \epsilon \cdot g_{t+1})\odot \bm{M}_{obj},
\end{equation}
\begin{equation}\label{eq:gtdefine}
    g_{t+1} = \mu\cdot g_t + \frac{\nabla_{\bm{M}}\mathcal{L}_{obj}(f(\bm{x}_{adv}), y;\bm{M}_t)}{\parallel \nabla_{\bm{M}}\mathcal{L}_{obj}(f(\bm{x}_{adv}), y;\bm{M}_t) \parallel_1}.
\end{equation}
 In practice, however, we find some elements in $\bm{M}_{t+1}$ output by Eq.(\ref{eq:gtdefine}) are still scattered. To smooth these elements, we fine-tune the gradient in the optimization process as follows: 
\begin{equation}
    g_{t+1} = g_{t+1}\odot Norm(\bm{M}_t * \mathcal{K}_{gau}),
    \label{eq:finetune1}
\end{equation}
where $Norm(\cdot)$ is normalization, $\mathcal{K}_{gau}$ is Gaussian kernel, $*$ is convolution. Eq.(\ref{eq:finetune1}) considers the elements' context information to remove outlier points, making the shape smooth.

\subsection{Analysis for Cover Image}
\label{sec: cover}
The infrared thermal imagery of an object is determined by the thermal radiation emitted by the object. Based on this feature, we put a patch of thermal insulation material at a specific location of the target to reduce the thermal radiation, thus changing the distribution of the infrared radiation of the object to fool the detector. 



\begin{figure}[h]
\begin{center}
\includegraphics[width=0.43\textwidth]{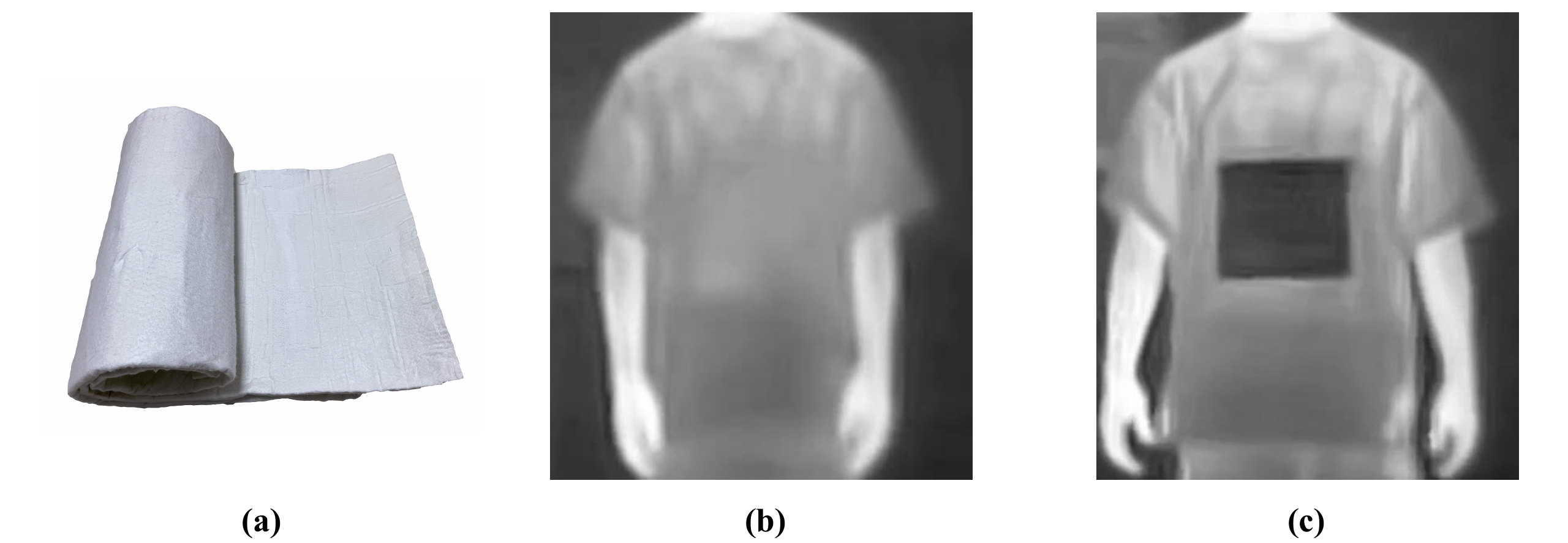}
\end{center}
\vspace{-0.5cm}
\caption{The comparison of infrared images taken by the infrared camera between (b) normal object and (c) object with patches made by the (a) aerogel material.}
\label{fig:radiation_real}
\end{figure}

In practice, we choose the aerogel material as the thermal insulation material for its low thermal radiation and low thermal emissivity. The material and its heat-insulating effect can be seen in Figure \ref{fig:radiation_real}. We can see the material on the object can shield most of the thermal radiation of the target object and reduce the thermal radiation of the corresponding area to a certain extent. We find that the gray values corresponding to the region with material in the infrared image are almost the same, so these gray values are set to  $\bm{\hat{x}}$.

\subsection{The Overall Algorithm}
\begin{figure*}[t]
\begin{center}
\includegraphics[width=0.85\textwidth]{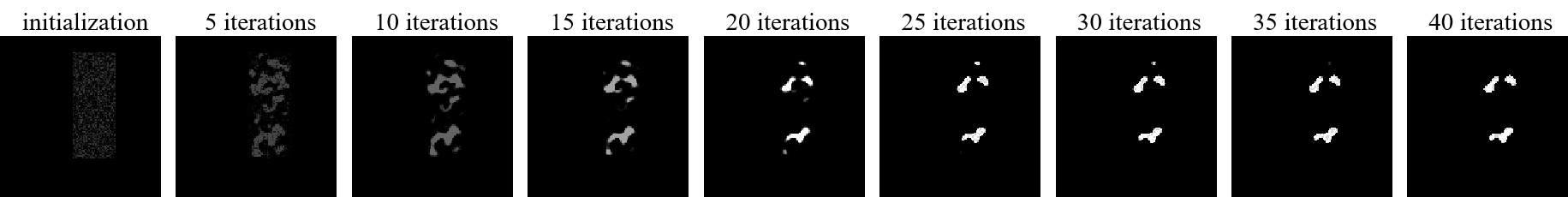}
\end{center}
\vspace{-0.5cm}
\caption{The optimization process of mask $\bm{M}$ for every 5 iterations, where we can see the mask gradually becomes aggregated and stable. }
\label{fig:mask_opt}
\vspace{-0.3cm}
\end{figure*}

The overall algorithm for generating adversarial infrared patches is summarized in Algorithm \ref{alg:orithm}. Given the pedestrian detector and infrared image, Algorithm \ref{alg:orithm} will output the adversarial infrared patch and adversarial image. The process of optimized mask $\bm{M}$ is given in Figure \ref{fig:mask_opt}. We see that the mask gradually converges to a stable shape and location. 

\begin{algorithm}[h]  
    \renewcommand{\algorithmicrequire}{\textbf{Input:}}
    \renewcommand{\algorithmicensure}{\textbf{Output:}}
    \caption{Optimization for Adversarial Infrared Patches}  
    \label{alg:orithm}
    \begin{algorithmic}[1]   
        \REQUIRE {Pedestrian detector $f(\cdot)$, clean infrared image $\bm{x}$, cover image $\bm{\hat{x}}$, restriction region $\bm{M}_{obj}$, max iteration $T$, max size $\varepsilon_{max}$, threshold $s_{thr}$.}
        \ENSURE Infrared patch $\bm{M}^{*}$, and adversarial example $\bm{x}_{adv}^{*}$
   
        \STATE {\small Initialize $\bm{M}_0$ randomly in $\left[0,1\right]$, $g_{0} = 0$.}
        \STATE {Restrict $\bm{M}_0$ in the  object region: $\bm{M}_0$ = $\bm{M}_0\odot \bm{M}_{obj}$}
        \FOR{$t = 0$ to $T$}  
            \STATE{Compute the gradient $g_{t+1}$ according to Eq.(\ref{eq:gtdefine})}
            \STATE{Finetune the gradient $g_{t+1}$ according to Eq.(\ref{eq:finetune1})}
            \STATE{Update the mask $\bm{M}_{t+1}$ according to Eq.(\ref{eq:upgr})}
            \STATE{
            Compute  $\bm{x}_{t+1}^{adv}$ according to Eq.(\ref{eq:advget})
            }
            \STATE{
            Perform the prediction $y=f(\bm{x}_{t+1}^{adv};\theta)$, and select the top-1 bounding box $(b_1,s_1)$
            }
            \IF{$s_1\leq s_{thr}$ and $\vert\vert\bm{M}_{t+1}\vert\vert_1\leq \varepsilon_{max}$}
                \STATE{$\bm{M}^{*}$ = $\bm{M}_{t+1}$; $\bm{x}_{adv}^{*}$ = $\bm{x}_{t+1}^{adv}$; break;}
        	\ENDIF
        \ENDFOR  
        \STATE{\small\textbf{return} $\bm{M}^{*}$, $\bm{x}_{adv}^{*}$}
  \end{algorithmic}  
\end{algorithm}

\section{Experiments}
In this section, we first perform the ablation study for our method in the digital world against pedestrian detector, and then give the attack performance in the physical world against pedestrian detector and vehicle detector, respectively. Finally, the result against defense is given.  More experiments can be found in the \emph{Supplemental Material}.

\subsection{Simulation of Physical Attacks}

\noindent{\textbf{Dataset}}: We use the FLIR ADAS dataset\footnote{https://www.flir.com/oem/adas/
adas-dataset-form/} to simulate the physical attacks. Similar to \cite{zhu2022infrared} and \cite{zhu2021fooling}, we select the images containing pedestrian as our dataset.  The training set includes 7873 images, and the test set includes 2027 images. We then choose the test images that  can be successfully detected by the target model with a high probability as the final images to perform attacks. Thus, the clean AP is 100\%.

\noindent{\textbf{Target detector}}: For pedestrian detection task, we also choose YOLOv3 \cite{redmon2018yolov3} based detector for its high speed, which is the same to \cite{zhu2022infrared} and \cite{zhu2021fooling}. We choose the officially pretrained weights as the initialized weights and then re-train the model on the FLIR ADAS training dataset. 

\noindent{\textbf{Metrics}}: Attack Success Rate (ASR) and Average Precision (AP) are used to evaluate the attack performance. ASR denotes the ratio of successfully attacked images out of all the test images. AP is computed by measuring the region under the Precision-Recall (PR) curve. 

\subsubsection{Comparisons with SOTA methods}
\begin{figure}[t]
\begin{center}
\includegraphics[width=0.95\linewidth]{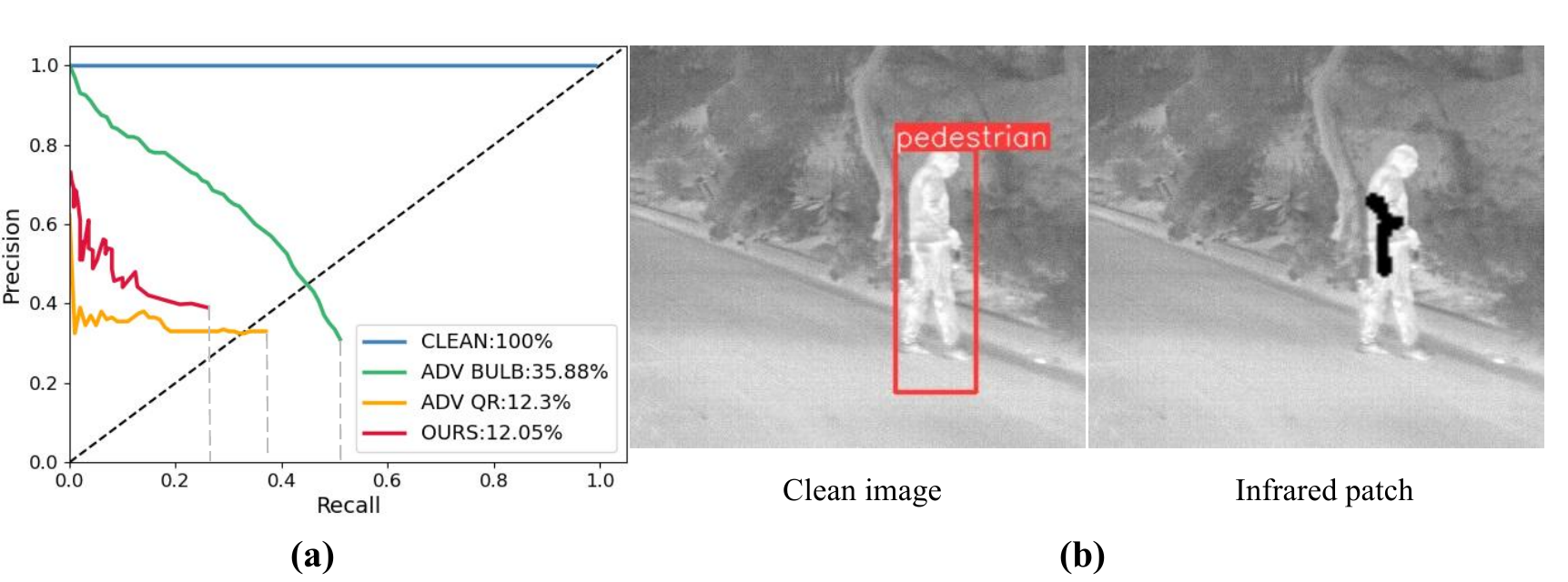}
\end{center}
\vspace{-0.5cm}
\caption{Comparisons with SOTA infrared attacks in  digital world.
}
\label{fig:digital1}
\end{figure}

We first test the attack performance of our method in the digital world, and compare with adversarial clothing \cite{zhu2022infrared} and adversarial bulbs\cite{zhu2021fooling} in the same dataset.  The Precision-Recall curves are plot in Figure \ref{fig:digital1} (a) and the corresponding AP values are also listed, where we see that the AP of pedestrian detector drops to 12.05\% from 100\% after the attack by our method, which is obviously lower than adversarial bulbs' 35.88\%, and slightly lower than adversarial clothing's 12.30\%. These results verify the advantage of our method. A qualitative example is given in Figure \ref{fig:digital1} (b). 

\begin{figure}[t]
\begin{center}
\includegraphics[width=0.9\linewidth]{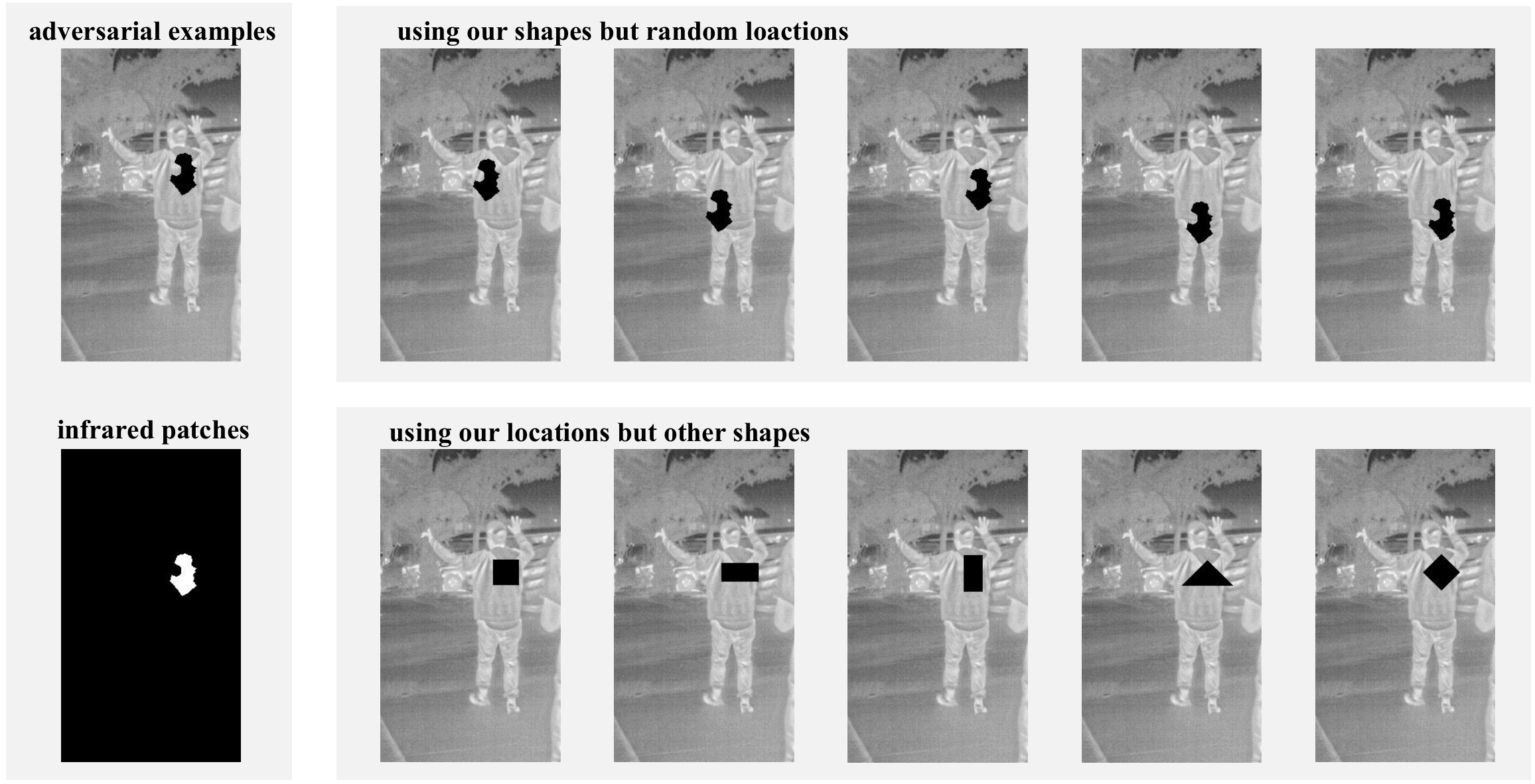}
\end{center}
\vspace{-0.5cm}
\caption{The experimental setting of ablation study for the infrared patches' shapes and locations.}
\label{fig:shape&location}
\vspace{-0.3cm}
\end{figure}

\begin{table}[h]
    \caption{Ablation study for infrared patches' shapes and locations.}
    \vspace{-0.4cm}
  \begin{center}
    \begin{tabular}{c|c|c}
    \hline
    Our location & Our shape &  ASR \\
    \hline 
      --     &--  &0\%    \\
    \checkmark   & & 54.16\%  \\
    & \checkmark &  59.68\%  \\
    \checkmark & \checkmark & \textbf{93.60\%}   \\
    \hline
    \end{tabular}
    \label{tab:trans1}
  \end{center}
  \vspace{-0.9cm}
\end{table}
\subsubsection{Effects of shapes and locations}
 To investigate the effects of different shapes and locations, we here give the ablation study. Specifically, we first generate adversarial infrared patches by solving Eq.(\ref{eq:totalloss}) to achieve the optimal shapes and locations (the left sub-figure in Figure \ref{fig:shape&location}) for a given pedestrian. Then, we conduct two additional experiments. One is fixing the optimal shapes while using random locations (the top row in the right sub-figure in Figure \ref{fig:shape&location}), then computing the average ASR under these five random locations. This setting tests the effects of optimal locations to the attacks.  The other is fixing the optimal locations while changing different patches' shapes (the bottom row in the right sub-figure in Figure \ref{fig:shape&location}). These shapes can be square, horizontal rectangle, vertical rectangle, triangle, rhombus, etc. Then we compute the average ASR under these five shapes.  This setting tests the effects of optimal shapes to the attacks. 
 The results are listed in Table \ref{tab:trans1}, where we can see (1) With the optimal shapes and locations, adversarial infrared patches achieve the best ASR (93.60\%). (2) When only using the optimal shapes or locations, there still exists the attacking ability for adversarial infrared patches, but the performance is weaker.  And the locations and shapes have a similar attacking ability (54.16\% vs 59.68\%).  (3) When neither our location nor our shape is used, the ASR decreases to 0\%. The contrast in Table \ref{tab:trans1} demonstrates the importance of patches' shapes and locations to the attack.

\begin{figure}[t]
\begin{center}
\includegraphics[width=0.35\textwidth]{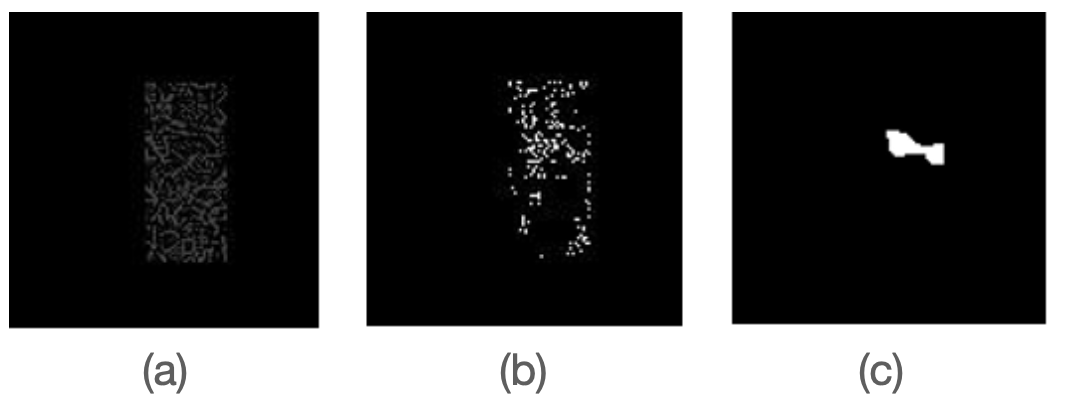}
\end{center}
\vspace{-0.6cm}
\caption{Example for infrared mask $\bm{M}$. (a) With only $\mathcal{L}_{attack}$. (b) With $\mathcal{L}_{attack}$ and $\mathcal{L}_{binary}$. (c) With $\mathcal{L}_{attack}$, $\mathcal{L}_{binary}$ and $\mathcal{L}_{agg}$. }
\label{fig:ablation}
\vspace{-0.3cm}
\end{figure}

\begin{table}[h]
    \caption{Ablation study for different loss functions.}
    \vspace{-0.4cm}
  \begin{center}
    \begin{tabular}{c|c|c}
    \hline
    Loss functions & ASR &  Aggregation  \\
    \hline 
      $\mathcal{L}_{attack}$     &100\%  &0.0057     \\
    $\mathcal{L}_{attack}+\mathcal{L}_{binary}$     &99.60\%   & 0.0427 \\
     $\mathcal{L}_{attack}+\mathcal{L}_{binary}+\mathcal{L}_{agg}$                      & 93.60\% &  0.4651  \\
    \hline
    \end{tabular}
    \label{tab:trans9}
  \end{center}
  \vspace{-0.8cm}
\end{table}

\begin{figure*}[t]
\begin{center}
\includegraphics[width=0.85\linewidth]{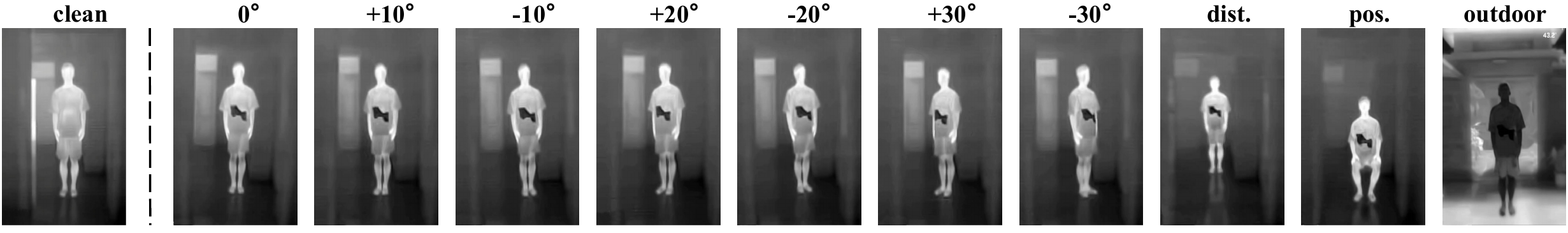}
\end{center}
\vspace{-0.4cm}
\caption{Visual examples of physical attacks with infrared patches under various angles, distances, postures, and scenes. 
}\vspace{-0.2cm}
\label{fig:situation}
\end{figure*}

\begin{table*}[h]
\caption{ASR in the physical world when changing distances, postures, angles, and scenes of pedestrian captured by infrared cameras.}
\vspace{-0.5cm}
  \begin{center}
    \begin{tabular}{c|c|c|c|c|c|c|c}
    \hline
    &  $0^{\circ}$ &  $\pm 10^{\circ}$ & $\pm 20^{\circ}$ & $\pm 30^{\circ}$ & dist. & pos. & outdoor \\
    \hline
    ASR & 94.67\% & 92.04\% & 89.05\% & 83.38\% & 77.25\% & 76.70\% & 87.70\%\\
    \hline
    \end{tabular}
    \label{tab:situations}
  \end{center}
\vspace{-0.3cm}
\end{table*}

\begin{figure}[h]
\begin{center}
\includegraphics[width=0.9\linewidth]{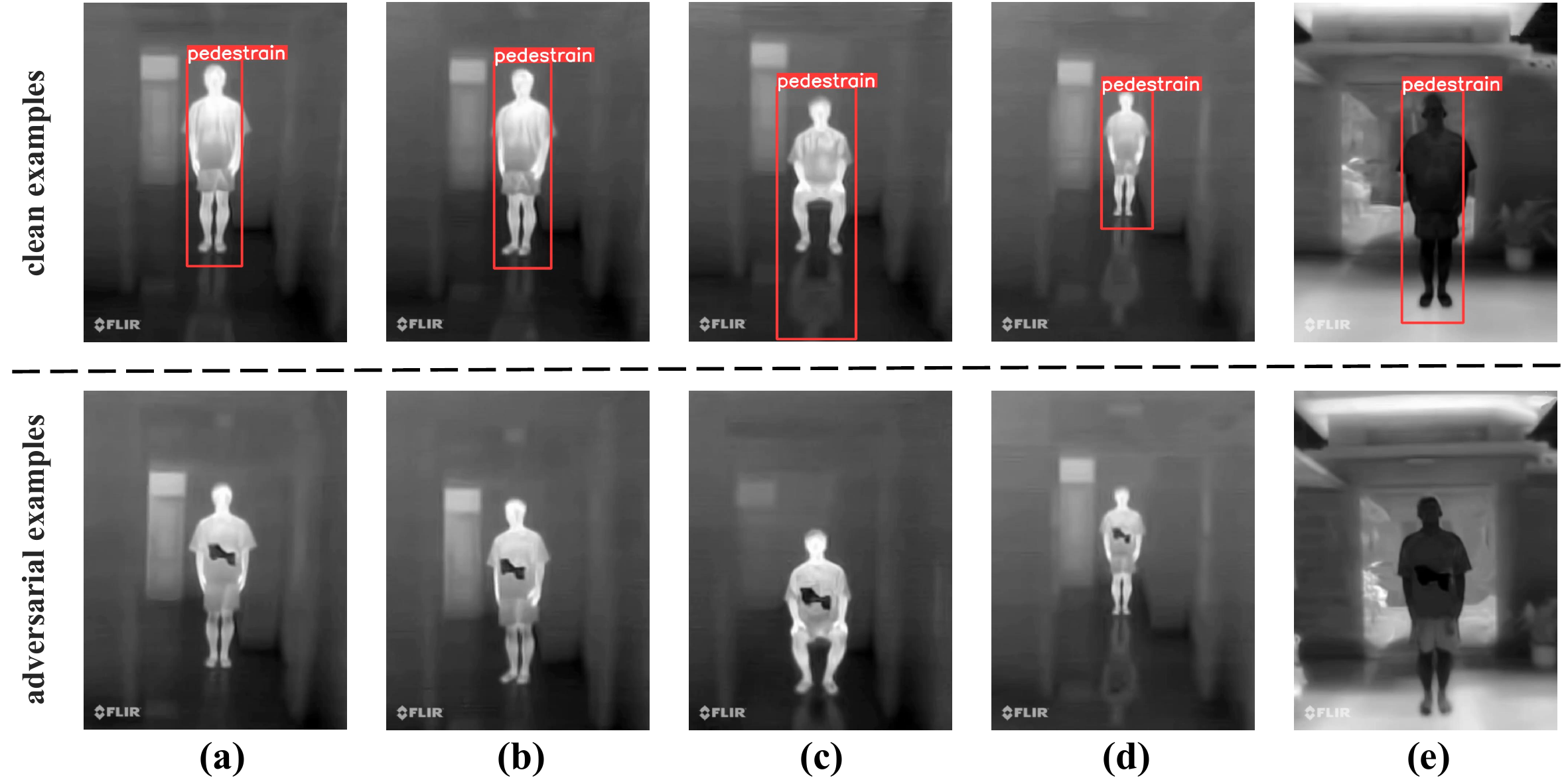}
\end{center}
\vspace{-0.5cm}
\caption{Visualization results of physical attacks with adversarial infrared patches. The top row lists the detection results on clean infrared images. And the bottom row lists the detection results on adversarial infrared images. We consider the front view in (a), different shooting angles in (b), sitting posture in (c), different shooting distances in (d), and outdoor scene in (e). }
\label{fig:phy}
\vspace{-0.5cm}
\end{figure}

\subsubsection{Effects of loss functions}

To visualize the impact of different loss functions, we step-wise add $\mathcal{L}_{attack}$, $\mathcal{L}_{binary}$, and $\mathcal{L}_{agg}$ to total loss as Eq.(\ref{eq:totalloss}) to optimize the infrared mask $\bm{M}$ and record the mask $\bm{M}$ after the optimization. An example of generated mask $\bm{M}$ on a pedestrian is given in Figure \ref{fig:ablation}. We can see that optimized with only $\mathcal{L}_{attack}$, the infrared mask $\bm{M}$ presents ``QR code" texture. When binary regularization $\mathcal{L}_{binary}$ is added, the elements in infrared mask $\bm{M}$ look like grainy dots, and more regions turn to black areas. It proves that the $\mathcal{L}_{binary}$ can effectively make the values of mask $\bm{M}$ get close to zero or one. Finally, when aggregation regularization $\mathcal{L}_{agg}$ is added, the grainy elements immediately gather together, and the infrared patch becomes  well-shaped on a specific location, which can be found that it is easy to be implemented in the real world. Besides, we also give the quantitative results of different loss functions in Table \ref{tab:trans9}, where we see with the integration of $\mathcal{L}_{binary}$ and $\mathcal{L}_{agg}$, the ASR has a slight reduction from 100\% to 93.60\%, but the aggregation degree shows a remarkable increase. A larger aggregation degree means a good physical implementation.

\subsection{Attacks in the Physical World}



\subsubsection{Physical attacks against pedestrian detector}
To test the performance of infrared patches under different physical conditions, we design various situations to conduct physical attacks and record the video to calculate the ASR. In default, we take videos at the 4 meters away from the standing pedestrian in the frontal view ($0^{\circ}$) within the indoor scene. For the angle problem, we change the angles with $\pm10^{\circ}$, $\pm20^{\circ}$, $\pm30^{\circ}$. For distance problem, we move the camera to 6 meters  from the default 4 meters. For posture problem, we change the pedestrian's posture from the default standing to sitting pose. For the scene problem, we change to outdoor from the default indoor. The visual examples of these situations are listed in Figure \ref{fig:situation}. During the shooting process, the pedestrian is asked to move their bodies within the range of $5^{\circ}$ of the current posture to take videos. For each situation, we shoot for 20 seconds at ten frames a second (about 200 frames in total) and compute the ASR versus the captured video. The pedestrian detection threshold is set as 0.5. Table \ref{tab:situations} lists the  quantitative results. It can be seen  that our infrared patches achieve a high ASR (94.67\%) in the frontal view. When the shooting angle changes, the ASR still maintains a high value (92.04\%, 89.05\%, and 83.38\%). When changing the distance from 4 meters to 6 meters, the ASR decreases to 77.25\%.  When changing the posture from standing to sitting, the ASR decreases to 76.70\%, and ASR decreases to 87.70\% when the scene is changed to outdoor. These results show that the impact of different shooting situations is relatively small to the infrared patches. In other words, as long as the shape of our infrared patch on the object can be completely captured by the camera,  the effects of adversarial attacks can be maintained. Some qualitative results before and after attaching the infrared patch on the pedestrian in different settings are given in Figure \ref{fig:phy}.

\begin{figure}[t]
\begin{center}
\includegraphics[width=0.9\linewidth]{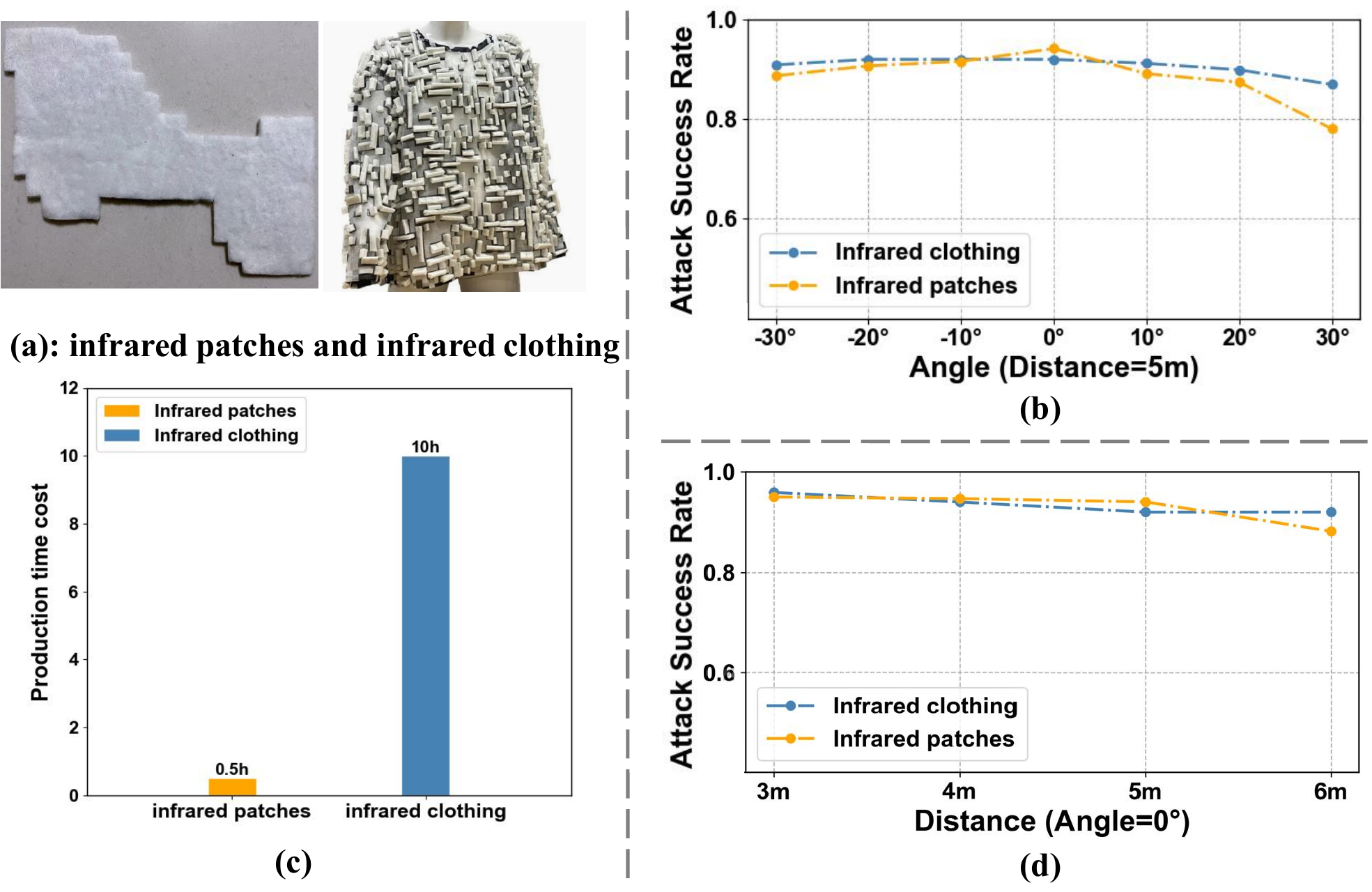}
\end{center}
\vspace{-0.5cm}
\caption{The comparisons of ASRs in real world at  different angles (b), different distances (d). We also report  the production time cost (c) of generating a physical adversarial example in real world between infrared clothing and infrared patches (a).}
\label{fig:comp}
\vspace{-0.3cm}
\end{figure}

We compare our infrared patches with the SOTA physical infrared attack: adversarial clothing \cite{zhu2022infrared}\footnote{We don't compare with adversarial bulbs \cite{zhu2021fooling} because they don't report corresponding physical results under different angles and distances, and reproducing the physical attacks is difficult owing to its poor implementation.}. We use the data provided in \cite{zhu2022infrared} and conduct our physical attack. For shooting angles, we select $0^{\circ}$, $\pm10^{\circ}$, $\pm20^{\circ}$, and $\pm30^{\circ}$ and fix the distance of 5 meters. For shooting distance, we select 3 meters, 4 meters, 5 meters, and 6 meters. We set the detection threshold as 0.7 like \cite{zhu2022infrared}. Finally, we compare the  time cost for infrared clothing and infrared patches in real world to evaluate their implementation difficulty of  physical attacks. The above results are in Figure \ref{fig:comp}. From the results, we can see that: (1) Our adversarial infrared patches have better performance than adversarial clothing when the pedestrian is captured in the frontal view, and show the competitive performance when the angle changes far away from the frontal view. Because our method can maintain the good attacks within $60^{\circ}$,  it is potential to extend it to achieve $360^{\circ}$ attack by attaching at least six infrared patches around the pedestrian.   (2) Our infrared patches have a similar performance with infrared clothing when the distance is located between the 3-th meter and 6-th meter, which shows the robustness against different distances. (3) The production time cost for infrared patches in real world (only about 0.5 hours) is far less than the infrared clothing (about 10 hours), which shows the production process of our infrared patches is much more convenient than infrared clothing. All in all, our adversarial infrared patches achieve the competitive attack performance in various situations but only cost five percent of their time to construct the physical attack compared with adversarial clothing attack, which demonstrates our advantage. 

\subsubsection{Physical attacks against vehicle detector}
Besides the pedestrian detection task, our adversarial infrared patch can also be applied to other tasks. In this section, we conduct the experiments of physical attacks against vehicle detection in the  aerial images \cite{razakarivony2016vehicle}. For that,  a two-stage object detector: Rotated Cascade R-CNN \cite{zhu2019rotated} is chosen as the target detector. In this way, we can verify our adversarial infrared patch against not only the one-stage YOLO detector above but also the two-stage R-CNN detector. Meanwhile, the experiments can also verify the wide application in both the horizontal object detector and rotated object detector.  Like the pedestrian detection task, we record the vehicle video captured in different angles and distances, and then compute the ratio of successfully attacked video frames out of all the frames as the ASR. Thus, we obtain a video with 750 frames and 696 frames are successfully attacked, leading to a 92.8\% ASR versus the vehicle detection. 

\begin{figure}[t]
\begin{center}
\includegraphics[width=0.95\linewidth]{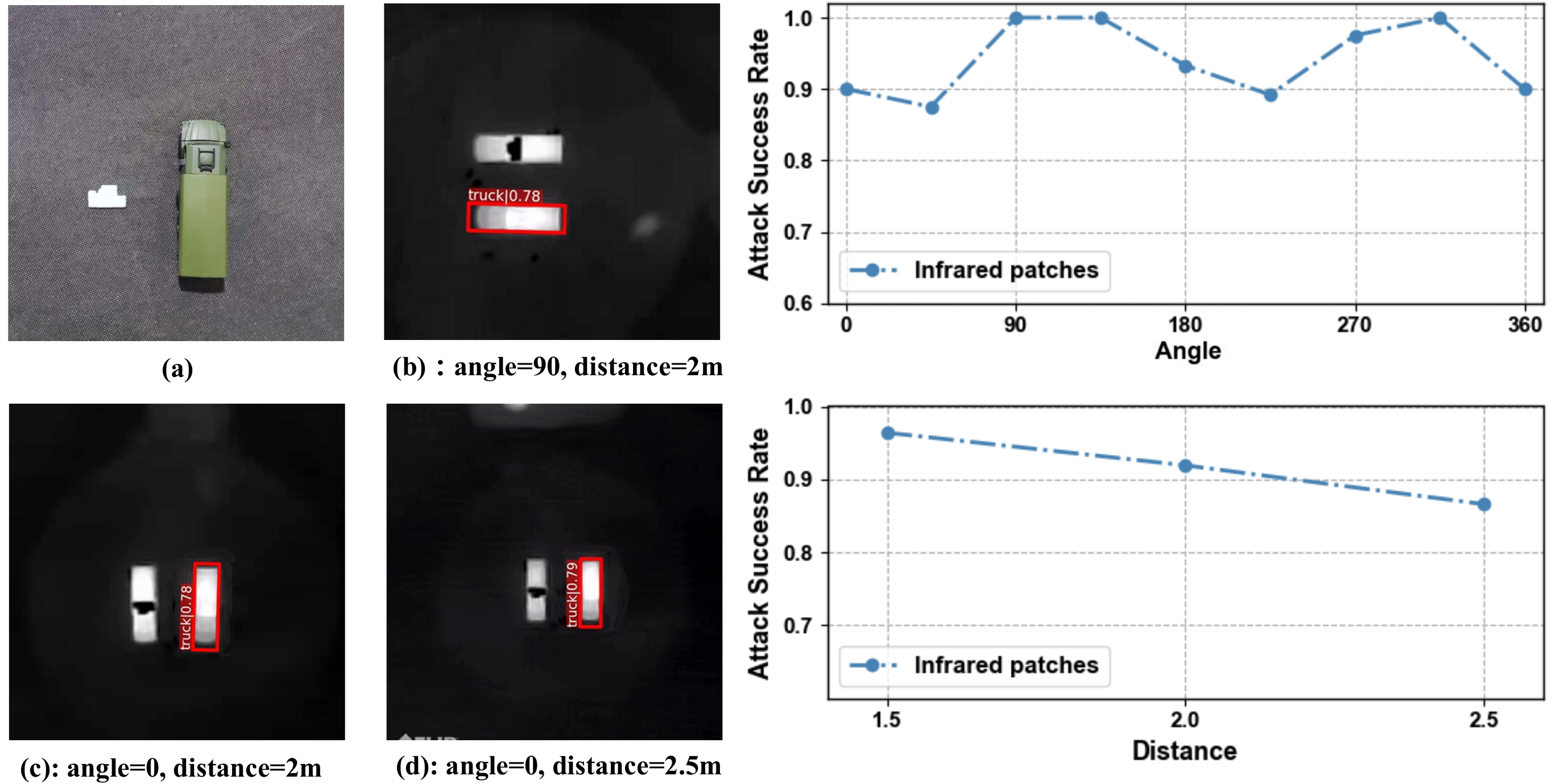}
\end{center}
\vspace{-0.5cm}
\caption{The qualitative and quantitative results of attacking  the vehicle detector in the physical world. }
\label{fig:extension}
\vspace{-0.4cm}
\end{figure}

The  results are given in Figure \ref{fig:extension}, where the left denotes the qualitative examples and the right denotes the quantitative ASR against different angles and distances.  To give a clear comparison, we list two vehicles in an image, where the vehicle attached by the infrared patch cannot be detected but the clean vehicle is successfully detected. Additionally, we see the patch' size is small versus the vehicle, but it indeed performs the successful adversarial attack, which can show the vulnerability of vehicle detector against the infrared patch. The right quantitative results show the high attack performance versus different angles and distances. It shows the similar trends with the pedestrian detection task. Because there are no published physical attacks against the vehicle detection task, we don't give the comparison here. 

\subsection{Defenses against Adversarial Infrared Patches}
We test three typical methods to defend our attack
method in the digital world. One is pre-processing defense: spatial smoothing \cite{xu2017feature},  the second one is adversarial training \cite{goodfellow2014explaining}, the third one is to randomize gradients \cite{qin2021random}. Results are given in Table \ref{tab:defense1}, where we see that (1) after spatial smoothing, ASR only drops 8\%. This is reasonable because the cover image in our method has the same value, the smooth operation cannot change the distribution.  (2) After randomizing gradients by adding random noises on the image, ASR is 14.8\% lower than no defense. (3) After adversarial training, ASR drops 29\%, which is still acceptable. It shows the robustness of our adversarial infrared patches to defense methods.

\begin{table}[h]
\caption{Results against the defense methods.}
\vspace{-0.5cm}
  \begin{center}
    \begin{tabular}{c|c|c|c|c}
    \hline
    Defenses &  No Defense  & \cite{xu2017feature} & \cite{qin2021random}&\cite{goodfellow2014explaining} \\
    \hline
    ASR    &93.6\%   & 85.6\%  &78.8\%  &64.6\%       \\
    \hline
    \end{tabular}
    \label{tab:defense1}
  \end{center}
\vspace{-0.8cm}
\end{table}

\section{Conclusion}
In this paper, we proposed physically feasible adversarial infrared patches with learnable shapes and locations, which can be solved through a carefully designed gradient-based optimization method. We constructed the infrared patches with thermal insulation material. The production of infrared patches was simple and the physical attack can be easy to realize. It only needed 0.5 hours to construct the adversarial patch in the physical world, which significantly outperformed the SOTA infrared attack methods. Experiments on the pedestrian detection and vehicle detection verified the effectiveness of the proposed method.

\section*{Acknowledgment}
This work is supported by National Key R$\&$D Program of China (No.2020AAA0104002), National Natural Science Foundation of China (No.62076018).

{\small
\bibliographystyle{ieee_fullname}
\bibliography{egbib}
}

\end{document}